\documentclass{article}

\usepackage[compact]{titlesec}
\titlespacing{\section}{0pt}{2ex}{1ex}
\titlespacing{\subsection}{0pt}{1ex}{0ex}
\usepackage{amsmath}
\usepackage[utf8]{inputenc}
\usepackage{algorithm,algcompatible,amssymb,amsmath}
\renewcommand{\COMMENT}[2][.5\linewidth]{%
  \leavevmode\hfill\makebox[#1][l]{//~#2}}
\algnewcommand\algorithmicto{\textbf{to}}
\algnewcommand\RETURN{\State \textbf{return} }
\usepackage[noend]{algpseudocode}
\makeatletter
\def\BState{\State\hskip-\ALG@thistlm}
\makeatother
\usepackage{tablefootnote}

\usepackage{PRIMEarxiv}

\usepackage[utf8]{inputenc} 
\usepackage[T1]{fontenc}    
\usepackage{hyperref}       
\usepackage{url}            
\usepackage{booktabs}       
\usepackage{amsfonts}       
\usepackage{nicefrac}       
\usepackage{microtype}      
\usepackage{lipsum}
\usepackage{fancyhdr}       
\usepackage{graphicx}       
\graphicspath{{media/}}     

\title{Leveraging Linear Independence of Component Classifiers: Optimizing Size and Prediction Accuracy for Online Ensembles

}
\newcommand{\orcid}[1]{\href{https://orcid.org/#1}{\includegraphics[width=8pt]{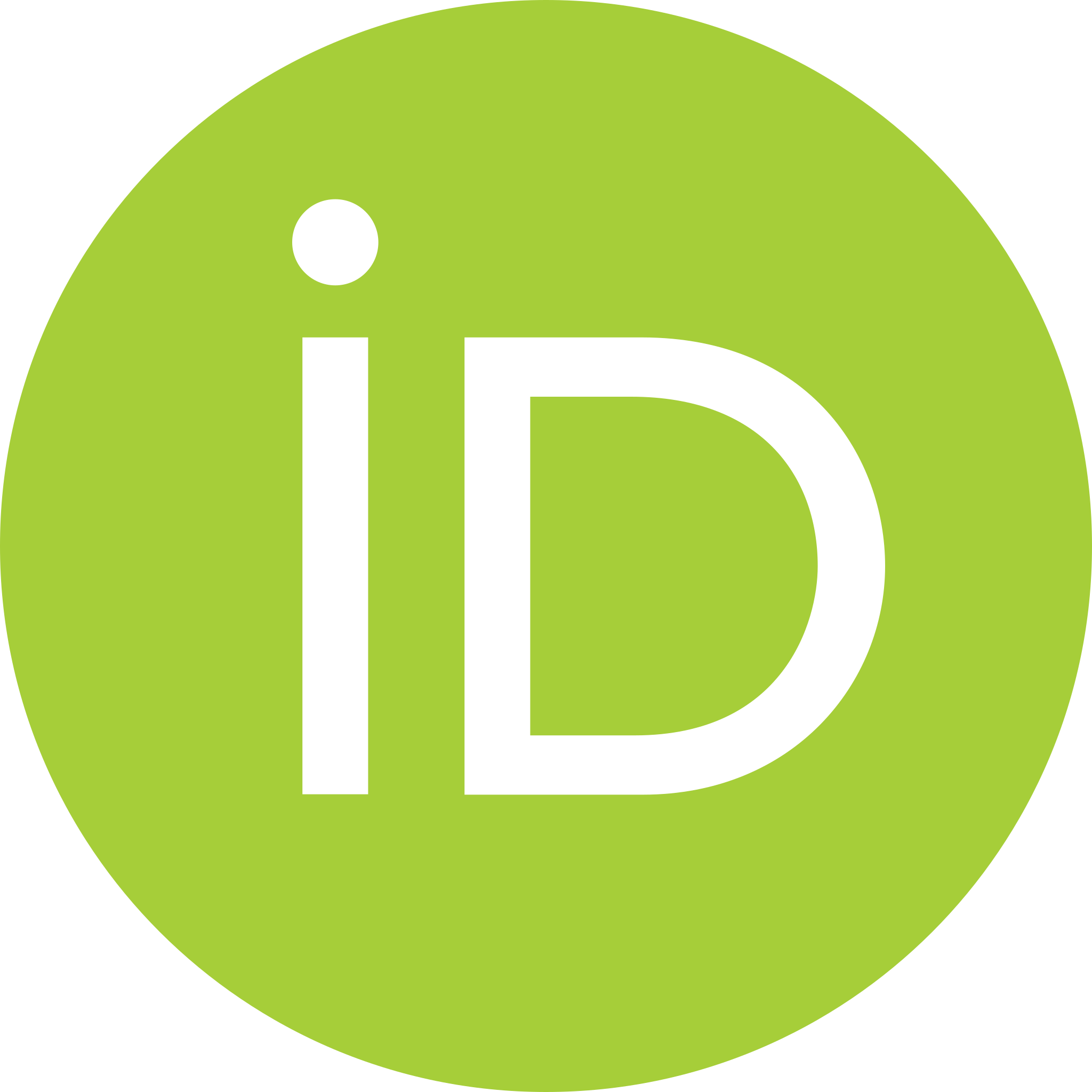}}}

\author{
  Enes Bektas \orcid{0009-0007-5186-8634} \\
  Bilkent Information Retrieval Group  \\
  Bilkent University \\
  Ankara\\
  \texttt{enes.bektas@ug.bilkent.edu.tr} \\
   \And
  Fazli Can \orcid{0000-0003-0016-4278} \\
  Bilkent Information Retrieval Group \\
  Bilkent University \\
  Ankara\\
  \texttt{canf@cs.bilkent.edu.tr} \\
}

\begin{document}
\maketitle

\begin{abstract}
Ensembles, which employ a set of classifiers to enhance classification accuracy collectively, are crucial in the era of big data. However, although there is general agreement that the relation between ensemble size and its prediction accuracy, the exact nature of this relationship is still unknown. We introduce a novel perspective, rooted in the linear independence of classifier's votes, to analyze the interplay between ensemble size and prediction accuracy. This framework reveals a theoretical link, consequently proposing an ensemble size based on this relationship. Our study builds upon a geometric framework and develops a series of theorems. These theorems clarify the role of linear dependency in crafting ensembles. We present a method to determine the minimum ensemble size required to ensure a target probability of linearly independent votes among component classifiers. Incorporating real and synthetic datasets, our empirical results demonstrate a trend: increasing the number of classifiers enhances accuracy, as predicted by our theoretical insights. However, we also identify a point of diminishing returns, beyond which additional classifiers provide diminishing improvements in accuracy. Surprisingly, the calculated ideal ensemble size deviates from empirical results for certain datasets, emphasizing the influence of other factors. This study opens avenues for deeper investigations into the complex dynamics governing ensemble design and offers guidance for constructing efficient and effective ensembles in practical scenarios.

\end{abstract}


\section{Introduction}
In the era of big data, where the volume and complexity of information have reached unprecedented levels, the need for highly accurate classification models gave rise to ensemble methods. As data generated by technological devices continues to exceedingly grow, relying solely on individual classifiers to classify vast amounts of data has diminished effectiveness. Ensembles, which use a set of classifiers to create a greater classifier that is better than each individual component classifier in the ensemble, offer a solution by using the collective intelligence of multiple models \cite{survey1, survey2, survey3, EnsembleMethodsinMachineLearning, errorReduction}. The ensemble methods are used in several areas including \textit{statistics}, \textit{machine learning}, \textit{pattern recognition} and \textit{knowledge discovery in databases} \cite{ensembleSelection}.

The two most used methods for ensemble creation are weighting and meta-combination \cite{ensembleBasedClassiers}. Weighting methods assign a weight to each classifier, which determines the strength of that classifier in the ensemble. Meta-combination techniques utilize meta-learning, where classifiers learn from training data and their own classifications on training data to improve prediction accuracy \cite{survey1, ensembleBasedClassiers}.

One crucial aspect of building an ensemble is determining the appropriate number of classifiers to include. Even though utilizing more classifiers is the main idea of ensembles, increasing the ensemble size comes with costs in terms of memory and time requirements. Several studies have investigated how the ensemble size can be determined. Bax suggested that any odd number of classifiers may be optimal \cite{bax}. Oshiro et al. stated that adding more classifiers beyond a certain threshold did not yield significant improvements \cite{oshiro}. Latinne et al. used the McNemar test to limit the number of classifiers in an ensemble \cite{latinne}. Pietruczuk et al. determine ensemble size by deciding if a new classifier increases the accuracy rate for the whole data stream \cite{PIETRUCZUK}. Hernàndez-Lobato et al. uses statistics to determine ensemble size \cite{HERNANDEZLOBATO}. Hu uses rough sets theory and database operations for constructing ensemble \cite{roughSets}.

The contributions of this study are: (1) By examining the literature on ensemble size and focusing on ensembles using weighting methods, this article aims to explain the relationship between ensemble size and prediction accuracy, and also as an outcome, (2) suggests a method to determine the optimum number of classifiers in an ensemble. Preliminary work on this topic was presented in \cite{lessIsMore}, \cite{infoRetrieval}, \cite{datasetLevel}, \cite{humanactivityrecog}. \cite{lessismorecikm}. Section 2 introduces the preliminary work. We used 'performance' and 'prediction accuracy' interchangeably.

\section{Related Works}
This paper focuses on a specific weighting method based on a geometric framework introduced by Wu and Crestani for information retrieval systems \cite{infoRetrieval}. Bonab and Can apply this framework to ensemble classification using weighted majority voting, where each classifier's vote is treated as a vector \cite{lessIsMore}. Wu and Ding extend the framework to dataset-level classification, further exploring its applicability \cite{datasetLevel}. Ding et al. apply this framework for human activity recognition \cite{humanactivityrecog}. 

Although this paper depends on a specific ensemble creation method, the outcomes can also be generalized.

\begin{table}
\caption{Definition of Symbols}
  \label{tab:freq}
  \begin{tabular}{|l p{11cm} |}
    \toprule
    Symbol & Definition\\
    \midrule
    D = \{$I_{1}, I_{2}, ...$\} & The data stream\\
    $I_k$ & An instance of data stream\\
    m & Number of class labels of data stream\\
    E = \{$C_{1}, C_{2}, ..., C_{n}$\} & The ensemble with n-number of component classifiers \\
    n & Number of classifiers in the ensemble\\ 
    $S_{i}$ & Vote vector of $i^{th}$ classifier of the ensemble\\
    $S_{ij}$ & $j^{th}$ component of $i^{th}$ classifier's vote\\
    $W_{i}$ & Weight of $i^{th}$ classifier in the ensemble\\
    $o$ & The ideal vector which is in the form (0,0,...,0,1,0,0,...,0)\\
    $V = W_{1}S_{1}+W_{2}S_{2}+...+W_{n}S_{n}$ & Vote of the ensemble\\
    $p_{k}$ & Probability of a classifier to be linearly dependent on the previous k-dimensional space\\
  \bottomrule
\end{tabular}
\end{table}
\raggedbottom

Let there be a data stream D with m-number of class labels and an ensemble E with n-number of component classifiers. In this geometric framework, each vote of a classifier is treated as a vector in m-dimensional space. Each vote is normalized so that the sum of their components is 1, i.e., $\sum_{j = 0}^{m-1}(S_{ij}) = 1$. An ideal vector, $o_k$, is defined as the vector which points to the true class label of an instance $I_k$. The ensemble's vote is a linear combination of its component classifiers. The loss function is defined as the Euclidian distance between the ideal vector and the ensemble's vote. The weights are calculated in order to minimize this loss function \cite{infoRetrieval, lessIsMore}. 

For weight assignment, Bonab and Can suggest that the number of classifiers should be equal to the number of class labels (n = m) \cite{lessIsMore, lessismorecikm}. However, Wu and Crestani proves that adding more classifiers to the ensemble generally improves performance; even in the worst case, it performs the same as before \cite{infoRetrieval}. These contrasting findings highlight the need for further investigation into the optimal ensemble size and the impact of adding more classifiers.

\section{A Basis to Understand Effect of Linear Dependency}

For weight calculation in this framework, we demonstrate that the linear independence of classifiers in the ensemble plays an important role because:
\begin{itemize}
    \item In this geometric framework, the ensemble is created by a linear combination of its component classifiers.
    \item Linear dependency of the votes of component classifiers determines the span of the ensemble's vote.
\end{itemize}
Theorem 1 demonstrates the importance of the linear independence of component classifiers.

\textit{Theorem 1:} If there are m-number of linearly independent votes for an instance $I_k$ of data stream D, then there exists unique $W_{1}, W_{2},..., W_{n}$ with $\sum_{i=1}^{n} W_i = 1$ such that the ensemble vote is equal to the ideal vector, i.e., V = o.

\textit{Proof:} Let the number of component classifiers be m (n = m), and all of them give linearly independent votes for the instance $I_k$. Then there exists a unique solution to the following equation.

\begin{gather}
    \begin{bmatrix}
        S_{11} & S_{21} & --- & S_{n1} \\
        S_{12} & S_{22} & --- & S_{n2} \\
        S_{13} & S_{23} & --- & | \\
        | & | & --- & | \\
        | & | & --- & | \\
        S_{1n-1} & S_{2n-1} & --- & S_{nn-1} \\
        S_{1n} & S_{2n} & --- & S_{nn}
    \end{bmatrix}
    \begin{bmatrix}
        W_{1} \\
        W_{2} \\
        W_{3} \\
        | \\
        | \\
        W_{n-1} \\
        W_{n}
    \end{bmatrix}
    =
    \begin{bmatrix}
        0 \\
        | \\
        0 \\
        1 \\
        0 \\
        | \\
        0 \\
    \end{bmatrix}
    = o
\end{gather}

It follows that

\[W_{1}S_{11} + W_{2}S_{21} + ... + W_{1}S_{11} = 0 \]
\[... + ... + ... + ...      = 0 \]
\[W_{1}S_{1k} + W_{2}S_{2k} + ... + W_{1}S_{1k} = 1 \]
\[... + ... + ... + ...      = 0 \]
\[W_{1}S_{1n} + W_{2}S_{2n} + ... + W_{1}S_{1n} = 0 \]

Sum all the rows and group them according to weights
\[W_{1}(S_{11} +S_{12} + ... + S_{1n}) + ... + W_{n}(S_{n1} +S_{n2} + ... + S_{nn})= 1 \]
Since $\sum_{i=1}^{n} S_{ki} = 1$ for all k between 1 and n, it follows that
\[W_{1} + ... + W_{n}= 1 \]

If there are more than m-number of classifiers (n > m) and m-number of them gives linearly independent votes, these m classifiers can be assigned to the ideal weights, and other n-m classifiers can be assigned to 0 weight.

\textit{Discussion: } This theorem shows that having an m-number of linearly independent votes among n votes in for an instance $I_k$ is enough to assign ideal weights.

For one instance $I_k$, only the m-number of classifiers may be enough. Nevertheless, since the votes given by classifiers change according to the instance $I_k$ of data stream D, the linear dependence of the classifier's votes changes. We utilize the probability of linear dependency of component classifiers to calculate the probability of having m-number of linearly independent votes among n component classifiers.

\textit{Definition: }Let $p_k$ denote the probability that a vote of a classifier is linearly dependent to a space with k dimensions. Assume that $p_k$s are the same for all classifiers.

\textit{Theorem 2: } Given the probabilities $p_{1}, ..., p_{m-1}$ the probability of having m linearly independent votes in an ensemble with n classifiers (n$\ge$m) is:

\begin{equation}
(\prod_{i=1}^{m-1}(1-p_{i}) ) (\sum_{k = 0}^{n - m}(\sum_{x_1+ ... + x_{m-1} = k} (\prod_{j=1}^{m-1}(p_{j}^{x_j}) ) )
\end{equation}

where each $x_i$ is a natural number between 0 and k.

\textit{Proof: } For every vote provided by classifiers, we construct a new set by adding one vote at a time and calculate the number of linearly independent vectors (this number is the same as the dimension of the space created by these votes) in this new set.

For the ensemble to have m number of linearly independent votes, the dimension of the new set should increase m-1 times, which has the probability $\prod_{i=1}^{m-1}(1-p_{i})$. An m-number of linearly independent votes can be achieved with at least m votes. However, if it cannot be achieved with m votes, some vectors  must have not increased the dimension with the probability of one of the $p_k$s. For example, the probability of having an m-dimensional system with m+1 vectors is (since all of them are independent events):
\[ (p_{1}+p_{2}+...+p_{m-1})\prod_{i=1}^{m-1}(1-p_{i}) \]    
By similar logic, the probability of having an m-dimensional system with m+2 vectors is:
\[ (p_{1}^2+p_{1}p_{2}+...+p_{m-2}p_{m-1}+p_{m-1}^2)\prod_{i=1}^{m-1}(1-p_{i}) \] 
By summing these probabilities up to n, the total probability of having an m-number of linearly independent votes among the n-number of classifiers is:
\begin{displaymath}
    (\prod_{i=1}^{m-1}(1-p_{i}) ) (\sum_{k = 0}^{n - m}(\sum_{x_1+ ... + x_{m-1} = k} (\prod_{j=1}^{m-1}(p_{j}^{x_j}) ) )
\end{displaymath}

where each xi is a natural number between 0 and k.

\begin{figure}[h]
    \centering
    \includegraphics[width=11cm]{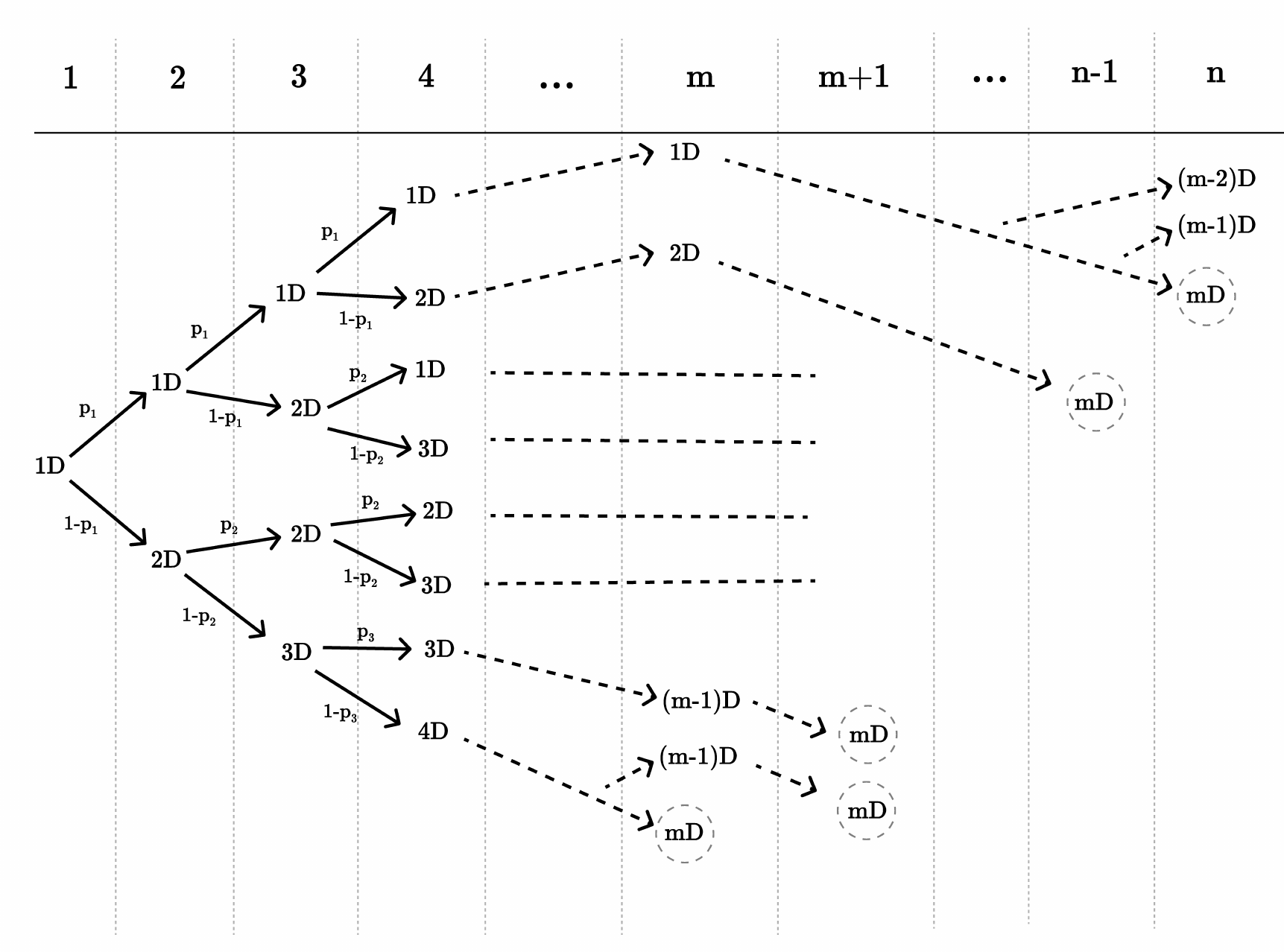}
    \caption{The probability tree for n-number of classifiers to have m-number of linearly independent votes. The numbers at the top of the figure represent the number of classifiers in the current system. D indicates dimension.}
\end{figure}

\textit{Discussion: } This theorem numerically shows how adding a new classifier to the ensemble affects the probability of having an m-number of linearly independent votes. However, we want this probability to be as high as possible so that we are sure there are enough classifiers. For this reason, its limit should be known.

\textit{Theorem 3: } If none of the $p_k$s is 1, then the probability of having m-number of linearly independent votes converges to 1 as n goes to infinity, i.e.

\begin{equation}
    \lim_{n \to +\infty}(\prod_{i=1}^{m-1}(1-p_{i}) ) (\sum_{k = 0}^{n - m}(\sum_{x_1+ ... + x_{m-1} = k} (\prod_{j=1}^{m-1}(p_{j}^{x_j}) ) ) = 1
\end{equation}

\textit{Proof: } There are three possible cases for the above formula:\newline
1) At least one of the $p_k$s is 1:\newline
    In this case the product $\prod_{i=1}^{m-1}(1-p_{i})$ evaluates to 0; hence the limit is 0. \newline
2) All of the $p_k$s are 0. \newline
    In this case both $\sum_{k = 0}^{n - m}(\sum_{x_1+ ... + x_{m-1} = k} (\prod_{j=1}^{m-1}(p_{j}^{x_j}) )$ and $\prod_{i=1}^{m-1}(1-p_{i})$ evaluates to 1; hence the limit is 1. \newline
3) None of the $p_k$ is 1, and at least one of them is not 0. \newline
    This case will be proven by induction.
    
    Base case: For m = 2, the equation (2) is equal to
    \[(1-p_1)(1+p_{1}+p_{1}^2 +p_{1}^3 + ...)\]
    Since $0<p_{1}<1$,
    \[1+p_{1}+p_{1}^2 +p_{1}^3 + ... = \sum_{i = 0}^{\infty}p_1 = \frac{1}{1-p_1}\]
    Hence,
    \[(1-p_1)(1+p_{1}+p_{1}^2 +p_{1}^3 + ...) = 1\]
    Let the theorem hold for m = k+1, i.e.,
    \[(1-p_1)...(1-p_k)(1+p_{1}+...+p_{k}+p_{1}^2+p_{1}p_{2}+...p_{k}^2+p_{1}^3+...) = 1\]
    \[\implies(1+p_{1}+...+p_{k}+p_{1}^2+p_{1}p_{2}+...p_{k}^2+p_{1}^3+...) = \frac{1}{(1-p_1)...(1-p_k)}\]
    
    Then it is enough to show that 
    \[1+p_{1}+...+p_{k+1}+p_{1}^2+p_{1}p_{2}+...p_{k+1}^2+p_{1}^3+... = \frac{1}{(1-p_1)...(1-p_{k+1})}\]
    Let 
    \[1+p_{1}+...+p_{k+1}+p_{1}^2+p_{1}p_{2}+...p_{k+1}^2+p_{1}^3+... = A\]
    Group the terms in the equation so that the terms with $p_{k+1}$ are gathered
    \[1+p_{1}+...+p_{k}+p_{1}^2+p_{1}p_{2}+...p_{k}^2+p_{1}^3+...\]
    \[+p_{k+1}+p_{k+1}p_{1}+...+p{k+1}^2+p_{k+1}p_{1}^2+...\]
    \[=1+p_{1}+...+p_{k}+p_{1}^2+p_{1}p_{2}+...p_{k}^2+p_{1}^3+...\]
    \[+p_{k+1}(1+p_{1}+...+p_{k+1}+p_{1}^2+p_{1}p_{2}+...p_{k+1}^2+p_{1}^3+...)\]
    \[=\frac{1}{(1-p_1)...(1-p_k)}+p_{k+1}(A)=A\]
    \[\implies A = \frac{1}{(1-p_1)...(1-p_{k+1})}\]
    Hence,
    \[(1-p_1)...(1-p_{k+1})(1+p_{1}+...+p_{k+1}+p_{1}^2+p_{1}p_{2}+...p_{k+1}^2+p_{1}^3+...) = 1\]

\textit{Discussion: } Theorem 2 showed how adding new classifiers affects the probability of having an m-number of linearly independent vectors. Theorem 3 shows that this probability converges to 1 when n gets larger and larger.

\begin{table}[ht]
\caption[specTable]{Dataset Specifications. The first half is synthetic, the second half is real-life datasets. The calculated number of classifiers with the equation in Theorem 2 is given as Ideal \#Classifiers \footnotemark}
\centering
\begin{tabular}{| l r r r r |}
 \hline
 Dataset & \#Inst. & \#Attr. & \#Class Labels & Ideal \#Classifiers \\ [0.5ex] 
 \hline\hline
 RBF2 & 100,000 & 20 & 2 & 5 \\ 
 \hline
 RBF4 & 100,000 & 20 & 4 & 11 \\
 \hline
 RBF8 & 100,000 & 20 & 8 &  47\\
 \hline
 RBF16 & 100,000 & 20 & 8 &  274\\
 \hline
 Airlines & 539,383 & 7 & 2 &  4\\
 \hline
 Elec & 45,312 & 6 & 2 & 6\\ 
 \hline
 Rialto & 82,250 & 27 & 10 & 31\\ [1ex] 
 \hline
  Covtype & 581,012 & 54 & 7 &  93\\
 \hline
\end{tabular}
\end{table}

\footnotetext{Real-life datasets can be found at: https://anonymous.4open.science/r/concept-drift-datasets-scikit-multiflow-1261/README.md}

These three theorems explain the relationship between the number of component classifiers and ensemble performance. Furthermore, thinking backward for Theorem 2, we can set a target probability of having m-number of linearly independent votes, and calculate the minimum number of component classifiers n to reach that probability. This method provides us with a suggestion for the ensemble size.

\section{Experiments}
\subsection{Setup}
The experiments are conducted to investigate the relationship between the number of classifiers and the accuracy rate. Additionally, we calculate the minimum number of classifiers required using the equation in Theorem 2 with a target probability set at 0.999. For experiments, we use GOOWE (Geometrically Optimum and Online-Weighted Ensemble) as it already implements the geometric framework introduced \cite{goowe}. Hoeffding Tree is used as the base classifier of GOOWE \cite{hoeffding}.

To evaluate the performance of the ensemble, we employ four real-world and four synthetic datasets. The synthetic datasets are created using the random RBF (Radial Basis Function) generator available in the scikit-multiflow library of Python. The specific details of the datasets are given in Table 2.

The probabilities $p_k$s (k=1,...,m-1) are experimentally calculated via Algorithm 1. We extend the GOOWE algorithm so that it also calculates these probabilities when iterating each vote.

\begin{algorithm}
\caption[algorithm]{Algorithm for Calculating p Values\footnotemark}\label{euclid}
\begin{algorithmic}[1]
\State $p\_array \gets \text{zero array with length of \# class labels}$ \COMMENT{array that stores unsuccessful trials (vote is linearly }

\COMMENT{dependent) when adding new classifier vote to curMatrix.}
\State $p\_total \gets \text{zero array with length of \# class labels}$ \COMMENT{will be used to store total occurences of adding new  }

\COMMENT{vote to curMatrix for calculating p values.}
\For{each instance I in data stream}
    \State $voteMatrix \gets \text{votes of component classifiers of ensemble for I }$
    \State $curMatrix \gets \text{zero matrix with same dimensions as voteMatrix}$
    \State $prevDim \gets 1$
    \State $curDim \gets 1$
    \For{i in range(number of rows of voteMatrix)} \COMMENT{Add a row to curMatrix, and check the rank of it. } 
    
    \COMMENT{Update p\_array and p\_total accordingly.}
        \State $curMatrix[i] \gets voteMatrix[i]$
        \If{i == 0} 
            \State $continue$ \COMMENT{Since one vector is always linearly independent}
        \EndIf
        \State $prevDim \gets curDim$
        \State $curDim \gets \text{rank of curMatrix}$
        \If{curDim == precDim}
            \State $p\_array[curDim-1] \gets p\_array[curDim-1]+1$
        \EndIf
        \State $p\_total[prevDim-1] \gets p\_total[prevDim-1]+1$
        \If{curDim == length of p\_array}
            \State $break$
        \EndIf
    \EndFor
\EndFor
\For{i in range(length of p\_array)} \COMMENT{After this, p\_array[i] = $p_{i+1}$}
    \If{p\_total[i] == 0}
        \State $p\_array[i] \gets 1$
    \Else
        \State $p\_array[i] \gets p\_array[i]/p\_total[i]$
    \EndIf
\EndFor
\end{algorithmic}
\end{algorithm}

\footnotetext{For reproducability, the algorithms used for experiments can be found at: https://anonymous.4open.science/r/GOOWE-Extended-CF41/README.md}

\subsection{Results} Table 2 summarizes the specifications of the datasets used in the experiments, including the ideal number of classifiers required for each dataset. Figure 2 is the plot of Accuracy Rate vs. Ensemble Size for different datasets.

\begin{figure}[h]
    \centering
    \includegraphics[width = 12cm]{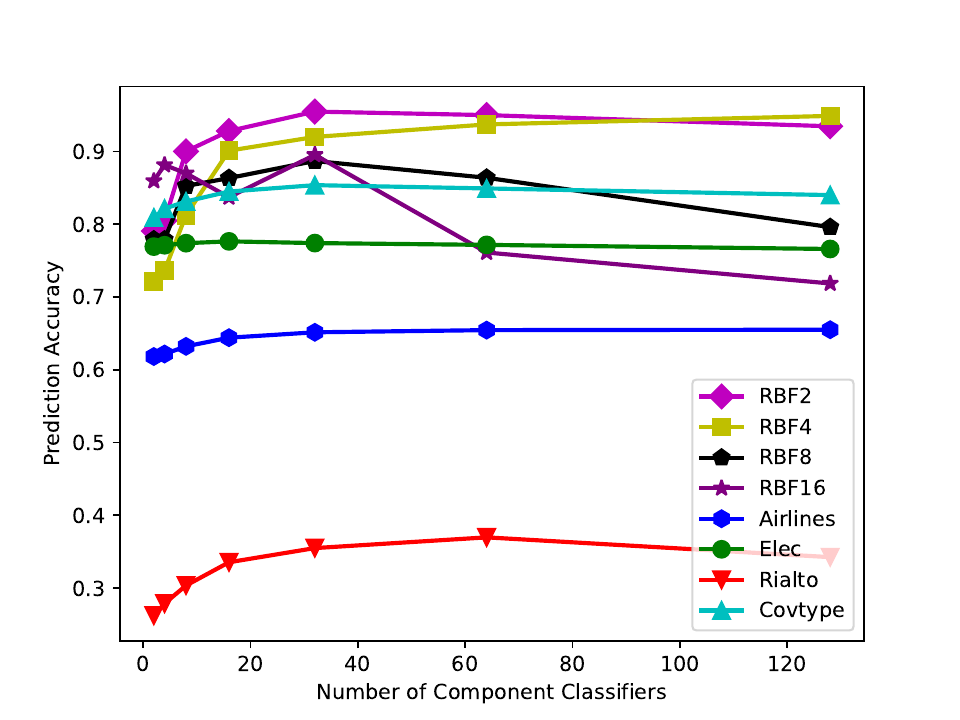}
    \caption{Prediction accuracy of the geometric framework with ensemble size ranging from 2 to 128. Synthetic datasets are named as RBF(number\_of\_class\_labels).}
\end{figure}

We made several observations during our experiments. Firstly,  Theorem 3 shows that equation (2) follows a convergent curve; the effect of adding more classifiers to the ensemble diminishes as the number of classifiers in the ensemble increases. This finding is in line with the rate of change of the accuracy rate, implying that beyond a certain point, adding more classifiers to the ensemble does not significantly improve performance. 

Furthermore, we confirmed that as the number of classifiers of an ensemble increases, the accuracy rate generally improves. This observation aligns with the trend predicted by Formula (3): as the number of classifiers increases, the probability of obtaining a sufficient number of linearly independent votes approaches 1. However, for most of the datasets, the accuracy rate decreases after some point, especially for 128 component classifiers. This shows that there are factors affecting the ensemble performance other than the linear independency of component classifiers that we want to explore. 

Moreover, for datasets with a low number of class labels, the calculated ideal ensemble size tends to be near the number of class labels. However, as the number of class labels increases, the ideal ensemble size deviates significantly from this linear relationship. This suggests that the complexity of achieving linearly independent votes among classifiers increases non-linearly with the number of class labels. Also, calculated ideal \#classifiers match with empirical results only for the RBF8 dataset. This again shows the importance of factors other than the linear independence of component classifiers.

\section{Conclusion \& Future Work}
In this paper, we show the importance of the linear dependence of classifiers for an ensemble. Using the probability of component classifiers being linear dependent, we provide a new perspective to understand how the number of classifiers in an ensemble affects the accuracy rate of that ensemble. Moreover, we provide a theoretical approach for determining ensemble size. The experiments we conducted show that, in general, the accuracy rate increases as the number of classifiers increases. This finding supports our theoretical framework.

We assumed that probabilities $p_k$s are the same for all classifiers in the ensemble, which means the classifiers are equally indistinguishable from each other. However, this is not true in real-life scenarios. In the future, we plan to investigate the individual classifier-level probability system to develop a more detailed theoretical framework.

\section*{Acknowledgments}
This study is partially supported by TÜBİTAK grant no. 122E271.

\bibliographystyle{unsrt}  
\bibliography{references}

\end{document}